\documentclass[journal]{IEEEtran}
\usepackage{graphicx}
\usepackage{titlesec}
\usepackage{float}
\usepackage{tabu}
\usepackage[hidelinks]{hyperref}
\usepackage{amsmath}
\usepackage{multirow}
\usepackage{array}
\usepackage{booktabs}
\usepackage{subcaption}
\usepackage{authblk}
\usepackage{comment}

\title{A Novel Global Context-aware Deep Neural Network for Enhanced Brain Tumor Segmentation using Magnetic Resonance Images}
\author{Sourjya Mukherjee, \IEEEmembership{Student Member, IEEE}, Ananya Bhattacharjee, \IEEEmembership{Student Member, IEEE}, R Murugan, \IEEEmembership{Member, IEEE}\thanks{This research was funded by the Ministry of Education,  Government of India as part of a Doctor of Philosophy degree at the National Institute of Technology Silchar.}
\thanks{The authors acknowledged NVIDIA Ltd for providing GPU under the NVIDIA Academic Grant.}
\thanks{This work did not involve human subjects or animals in its research.}\thanks{Sourjya Mukherjee, Ananya Bhattacharjee, and R Murugan are with Bio-Medical Imaging Laboratory, Department of Electronics and Communication Engineering, National Institute of Technology Silchar, Silchar, Assam, India, 788010. (Corresponding author e-mail: sourjyamukherjee261@gmail.com).}}

\begin{document}



\maketitle

\vspace{-20pt}
\begin{abstract}
 \noindent Brain cancer's severity necessitates precise brain tumor segmentation, which is crucial for effective brain tumor diagnosis. Manual identification, burdened by high costs, labor, and error risks, highlights the need for automated methods. In this study, we introduce the Global Context-aware Squeeze and Excite Residual UNet (GCSER-UNet), which facilitates a fusion of spatial and channel-wise attention and thus enhances the model's capacity to capture intricate spatial dependencies and contextual information. GCSER-UNet efficiently extracts tumor segments from multimodal MRI slices, delivering exceptional performance. Evaluations on benchmark databases exhibit its superiority, achieving a notable 94$\%$ dice score on the TCGA LGG dataset, surpassing the state-of-the-art dice score of 91.8$\%$. In the BraTS 2020 dataset, the proposed GCSER-UNet ensemble approach yielded dice scores of 95$\%$, 92$\%$, and 90$\%$ for the tumor regions—Whole Tumor (W), Tumor Core (T), and Enhancing Tumor (E), respectively. The current state-of-the-art dice scores were 94$\%$, 93$\%$, and 88$\%$. These compelling outcomes highlight the efficacy of GCSER-UNet in precise brain tumor segmentation and thus can aid neurologists in effective brain cancer management and treatment planning.
\end{abstract}
\def\abstractname{Impact statement}
\begin{abstract}

 \noindent The GCSER-UNet, our novel hybrid model, seamlessly incorporates U-Net, residual networks, Atrous Spatial Pyramidal Pooling (ASPP), and an innovative Global Context-aware Squeeze and Excite (GCSE) architecture in both encoder and decoder stages. Our intentional emphasis on simplicity prioritizes efficacy in response to the prevalent trend of architectures primarily tested on either high-grade gliomas or low-grade gliomas with elevated computational complexity. Critical structural enhancements to the standard UNet framework are pivotal for achieving consistent performance across glioma grades. Firstly, our observation that channels with heightened tumor-healthy tissue contrast exhibit higher standard deviation led to introducing the Global Context-aware Squeeze and Excite network. We replaced the global average pooling operation in the original Squeeze and Excite (SE) network with channel-wise mean and standard deviation for weighting. Secondly, spatial attention was incorporated into the channel-wise weighted feature maps. The result is a streamlined, effective solution demonstrating commendable performance across low-grade and high-grade gliomas. Thirdly, the ASPP module extracts multiscale features, and fourthly, an ensemble of three GCSER-UNets facilitates multiclass segmentation. This ensemble approach strategically manages computational complexity by employing a lightweight architecture for each tumor class, enabling optimal utilization of the model's capacity for class-specific learning. Remarkably, our architecture outperforms the original SE-Res-UNet and existing 3D approaches without relying on 3D context, underscoring its efficacy in advancing brain tumor segmentation. This streamlined yet potent approach aligns with our goal of providing a straightforward solution to the complexities associated with existing brain tumor segmentation architectures.
\end{abstract}


%
\IEEEpeerreviewmaketitle

\vspace{-7pt}
\section{Introduction}
\IEEEPARstart{B}{rain} tumor is an uncontrolled proliferation of somatic cells caused by a series of accumulating random mutations in critical genes that regulate cell growth and differentiation in the human brain \cite{r1-2}. Despite their rarity, brain tumors are highly lethal malignancies. The main body of the tumor can originate in the brain or manifest itself as a secondary growth with its primary origins in other organs \cite{r3}. Glioma, one of the most rapidly advancing types of brain tumor arising from glial cells in the brain, has been the focus of most recent brain tumor segmentation research. The typical survival duration for patients with glioblastoma, a particularly aggressive type of glioma, is fewer than 14 months \cite{r4}. Therefore, early detection of such malignancies can help implement a proper therapeutic regimen and more effective planning of surgeries.\cite{r4}.

Due to its superior soft tissue contrast and ubiquitous accessibility, magnetic resonance imaging (MRI) is widely considered the industry standard for detecting brain tumors \cite{r6}. However, precise segmentation is complex due to the features of brain tumors \cite{r6}. To improve the distinction between the various tumor subregions and between tumor and non-tumor tissue, data from a variety of complementary 3D MRI modalities, including T1, T1-ce (T1 with contrast agent), T2, and FLAIR (Fluid Attenuation Inversion Recovery), are combined. The T1-ce modality highlights the tumor boundary, while the T2 modality emphasizes the tumor area. FLAIR scans aid in distinguishing edema from cerebrospinal fluid (CSF) \cite{r7}. Integration of data from different MRI modalities can enhance the region-wise segmentation of brain tumors.

Manually segmenting MR scans is laborious and error-prone due to the significant variability in brain tumor location, size, and appearance \cite{r9-10}. Consequently, computer-assisted tumor segmentation has emerged as a highly sought-after solution. Automated segmentation methods enable the segmentation of brain tumors into different classes, such as necrotic tumor, enhancing tumor, tumor core, and edema, without requiring human intervention.

Convolutional Neural Networks (CNNs) have found extensive applications in biomedical imaging, particularly in segmentation tasks \cite{r8}. U-Net, a prominent CNN-based method, has been widely adopted for medical image segmentation \cite{ronneberger2015u}. Residual Networks (ResNets) help mitigate data loss during propagation by incorporating skip connections parallel to convolutional layers \cite{residual}. The squeeze-and-excitation (SE) network also introduces a content-aware mechanism that adaptively weighs each channel, enhancing the representation of essential features \cite{r18}. Atrous convolution, which captures multiscale data and enables precise feature resolution in deep convolutional neural networks, has also demonstrated significant utility \cite{aspp}. Motivated by the achievements of these networks, we propose the GCSER-UNet segmentation architecture.

This paper presents a deep neural network-based segmentation architecture encompassing substantial enhancements to the foundational U-Net architecture by employing an advanced Global Context-aware Squeeze and Excite (GCSE) mechanism. The proposed model is novel in two ways. First, it introduces an innovative ensemble model that amalgamates key elements from the UNet architecture, Residual Networks, Atrous Spatial Pyramidal Pooling (ASPP), and Global Context-aware Squeeze and Excite (GCSE) mechanism, thus resulting in a novel GCSER-UNet architecture. Notable structural refinements have been meticulously devised for both the contracting and expanding pathways through the seamless integration of the proposed GCSE mechanism. Second, the proposed model introduces an ensemble approach involving training three parallel models, each performing binary segmentation for the W, T, and E tumor classes. Given the extensive sample size, the binary segmentation for the three tumor classes is preferred over the multiclass approach to bolster the model's capacity for each class. Additionally, overlapping regions within the tumor classes render multiclass segmentation challenging. Binary segmentation allows focused delineation of unique class characteristics, ensuring precise segmentation while mitigating the complexities associated with inter-class boundaries, thus optimizing overall model performance. The proposed approach ensures precise delineation of each tumor class, facilitating accurate segmentation outcomes.

The main contributions of the proposed work are as follows:\\
\begin{enumerate}
\item A novel 2D CNN model based on the U-Net paradigm for the segmentation of brain tumors is implemented, which achieved better results than most 3D state-of-the-art
variants with a much lower computational cost. 

\item The novel GCSE blocks integrated post each Res-block module in the encoder and decoder networks dynamically recalibrate channel-wise features while seamlessly fusing channel-wise and spatial attention. This comprehensive approach overcomes the limitations of relying solely on channel-wise attention, significantly enhancing the model's capacity to capture intricate inter-channel dependencies and vital global context information. 

\item The ASPP applied at the bottleneck portion of the encoder enhances the receptive field of the proposed model, which significantly captures multi-scale information.\\
\end{enumerate}

The remainder of the paper is organized as follows. Section II is devoted to a survey of the literature. Section III delves into the materials and methods used in this study, including the model's construction and mathematical expressions. Section IV goes over the experiments and their results. The conclusion finishes Section V.
\section{Literature Review}
This section discusses several previous works that have provided valuable insights into developing the proposed model.

Ronneberger et al.\cite{ronneberger2015u} created the U-Net model for biological image segmentation based on the encoder-decoder paradigm. Sundaresan et al.\cite{r14} implemented a tri-planar architecture that consists of three two-dimensional UNet models, one used for each MR plane(coronal, sagittal, axial). This method yielded dice coefficient (DC) values of 83.8$\%$, 89.9$\%$, and 85.3$\%$ for the tumor subregions E, W, and T, respectively, on the BraTS 2020 dataset. Y. Xu et al. \cite{r15} incorporated a 3D ASPP module into a three-dimensional UNet and obtained DC values of 76.9$\%$, 87.1$\%$, and 77.9$\%$ for the tumor subcategories E, W, and T, respectively on the BraTS 2017 dataset. Varghese et al. \cite{r16} implemented a 23-layer deep fully connected 2D CNN based on the encoder-decoder framework and achieved DC values of 84.3$\%$, 84.1$\%$, and 77.3$\%$ for the tumor subcategories W, T, and, E, respectively on the BraTS 2017 dataset. Ding et al.\cite{stack_ding_2019} developed a  2D stacked UNet paradigm using the BraTS 2015 datasets and obtained DC values of 83$\%$, 67$\%$, and 59$\%$ for the tumor subcategories W, T, and E, respectively. Zhang et al.\cite{zhang_attention_2020} enhanced the performance of the UNet \cite{ronneberger2015u} by introducing residual building blocks to the original framework and adding gated attention units to the decoder. This approach yielded DC values of 87$\%$, 77$\%$, and 72$\%$ on tumor subcategories W, T, and E, respectively, on the BraTS 2017 dataset. Ilyas et al.\cite{ilyas_2022_hybrid} employed a novel weight alignment technique using attention modules with multiple dilation rates between the encoder and decoder skip links to promote enhanced feature mapping. On the BraTS 2018 dataset, this method produced DC values of 88$\%$, 76$\%$, and 65$\%$ for tumor subclasses W, T, and E, respectively. Ashraf et al.\cite{ottom2022znet} proposed the ZNet for the semantic segmentation of low-grade glioblastomas. The proposed framework was trained and evaluated on the TCGA LGG dataset, yielding a DC value of 91.5$\%$ on the testing data. A residual UNet created by Santosh et al.\cite{santosh2023resunet} was trained and evaluated on the TCGA LGG dataset, and the testing dataset produced a DC value of 90$\%$. Using a pre-trained VGG-16 backbone from the imagenet dataset and training data from the TCGA LGG dataset, Sourodip et al.\cite{r17} constructed a U-Net model and obtained a DC value of 91.6$\%$ on test data.\\
In summary, most of the works covered so far have been either trained and evaluated on datasets mostly of high-grade glioma (HG) volumes or exclusively on low-grade glioma (LG) slices. Due to their broad proliferation throughout healthy brain tissue, HGs are more accessible to segment than LGs. Also, class-based segmentation of tumor regions may prove to be quite tricky. An effective brain tumor segmentation model should function equally well on low-grade-glioma (LG) and high-grade-glioma (HG) volumes. To overcome this constraint, we propose a deep neural network-based semantic segmentation architecture showing promising segmentation results for both LGs and HGs. The proposed model was trained and validated independently using the BraTS 2020 and TCGA LGG datasets. 

\section{Materials and Methodology}
This section discusses in detail the datasets used, workflow, preprocessing techniques, and
the model architecture employed in sequential order.
\subsection{Materials}
\label{sec:materials}
In this study, two datasets have been used that are described in detail. 
\subsubsection{BraTS 2020 training dataset}
\begin{figure}[!h]
\setlength\tabcolsep{2pt}
\centering
\resizebox{0.45\textwidth}{!}{\begin{tabular}{c c c}

     \includegraphics[width=0.156\textwidth]{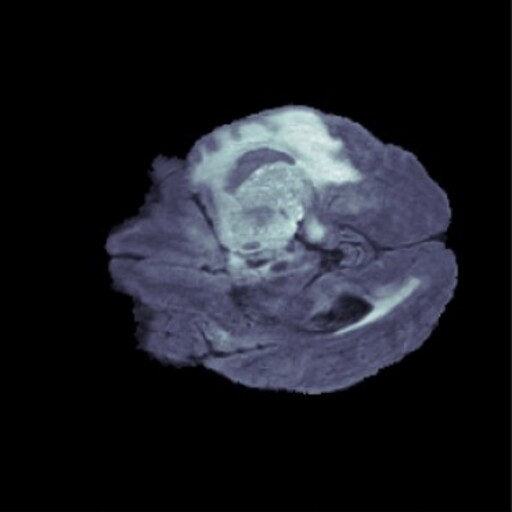} &
     \includegraphics[width=0.156\textwidth]{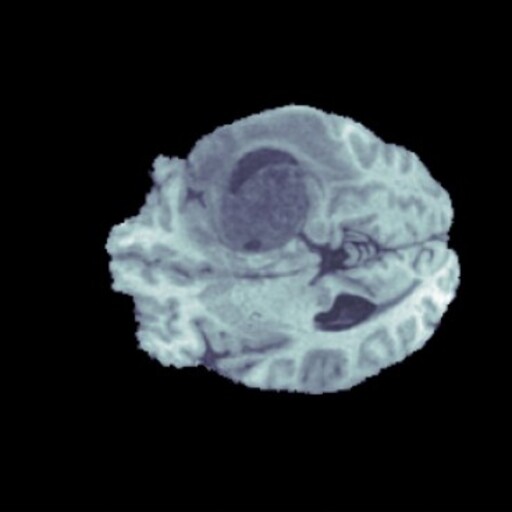} &
     \includegraphics[width=0.156\textwidth]{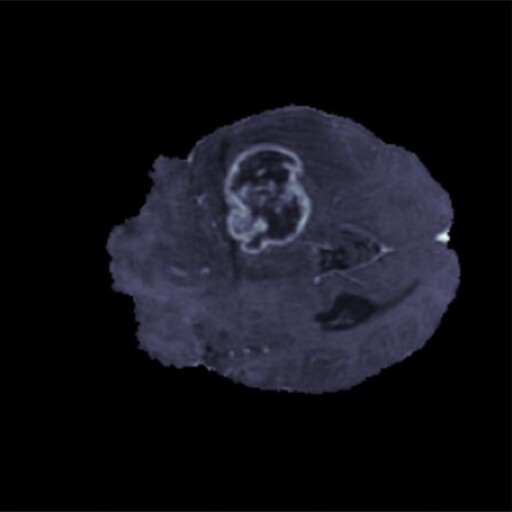}\\
     (a) & (b) & (c) \\
         
     \includegraphics[width=0.156\textwidth]{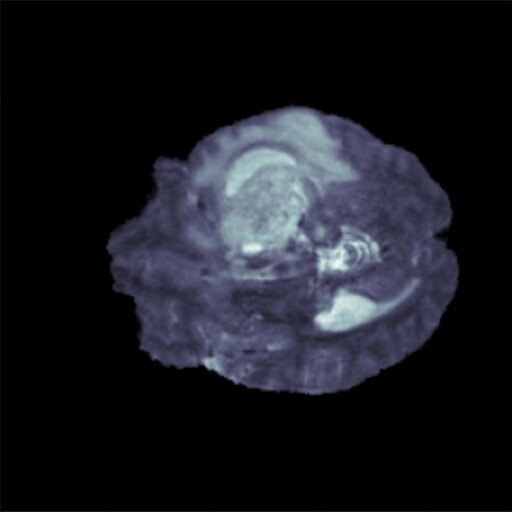} &
     \includegraphics[width=0.156\textwidth]{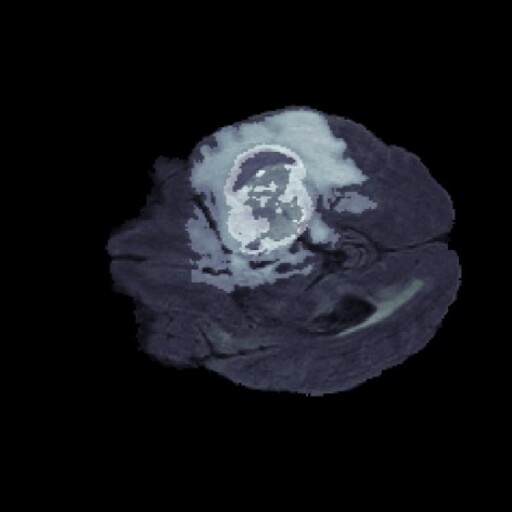} &
     \includegraphics[width=0.156\textwidth]{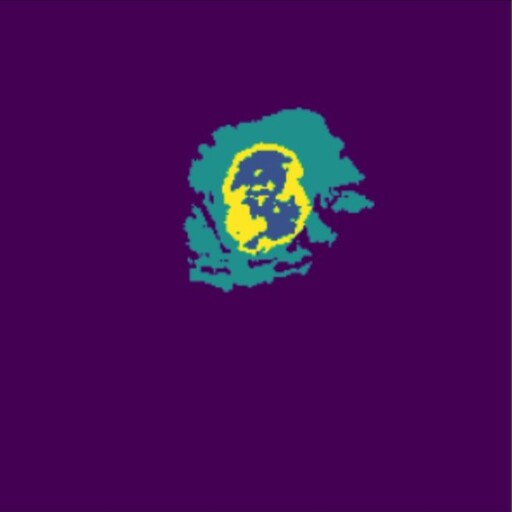}\\
     (d) & (e) & (f)

\end{tabular}}
\caption{Representative multimodal 2D slices from the BraTS 2020 training dataset and corresponding segmentation ground truth. (a) FLAIR modality, (b) T1 modality, (c) T1-ce modality, (d) T2 modality, (e) FLAIR modality superimposed with multiclass segmentation mask, (f) Multiclass segmentation mask.}
\label{fig:Brats_sample_images}
\end{figure}

\noindent The BraTS 2020 database \cite{brats_1,brats_2,brats_3} contains MRI volumes from 293 patients with HGs and 76 patients with LGs. T1, T1-enhanced (T1-ce), T2, and FLAIR MR volumes, along with physician segmentation outputs, were provided for each patient. The tumors have been manually categorized into three classes: edematous tissue, tissue having necrosis, and enhancing tumor region. Figure \ref{fig:Brats_sample_images} illustrates exemplary 2D multimodal MR slices from the dataset for patient volume no. 001, as well as the accompanying multiclass segmentation output produced by a physician..%
\subsubsection{TCGA LGG dataset}
\begin{figure}[!h]
\setlength\tabcolsep{2pt}
\centering
\resizebox{0.45\textwidth}{!}{\begin{tabular}{c c c}

     \includegraphics[width=0.156\textwidth]{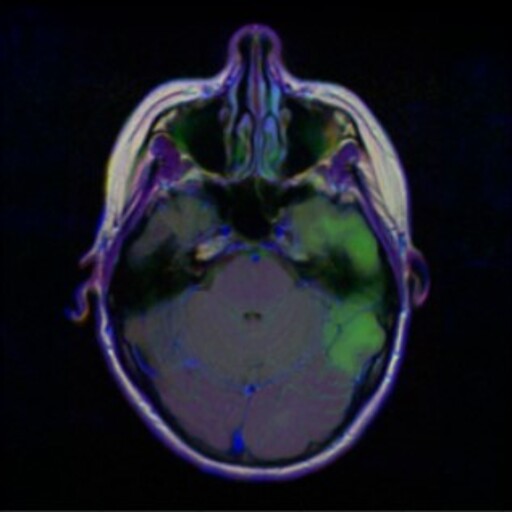} &
     \includegraphics[width=0.156\textwidth]{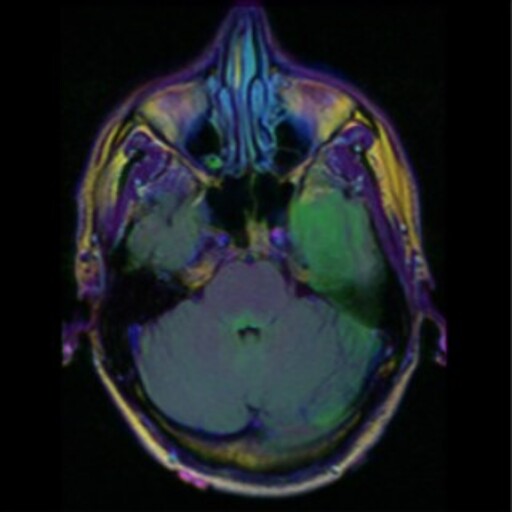} &
     \includegraphics[width=0.156\textwidth]{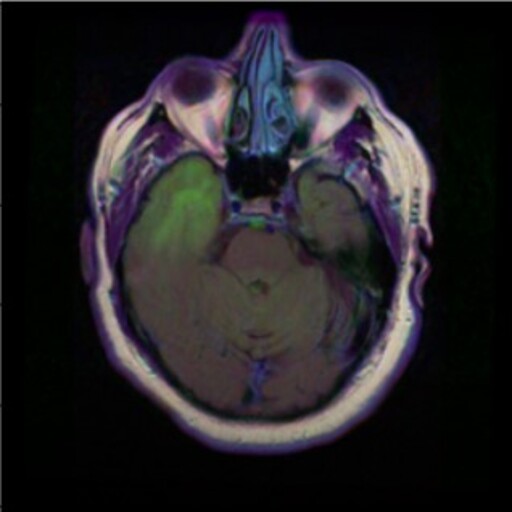}\\
     (a) & (b) & (c) \\
         
     \includegraphics[width=0.156\textwidth]{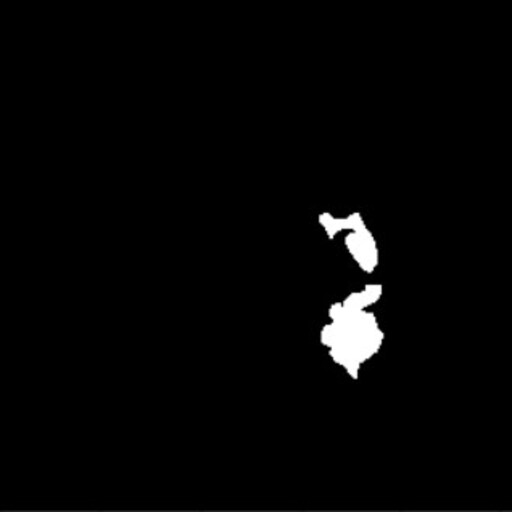} &
     \includegraphics[width=0.156\textwidth]{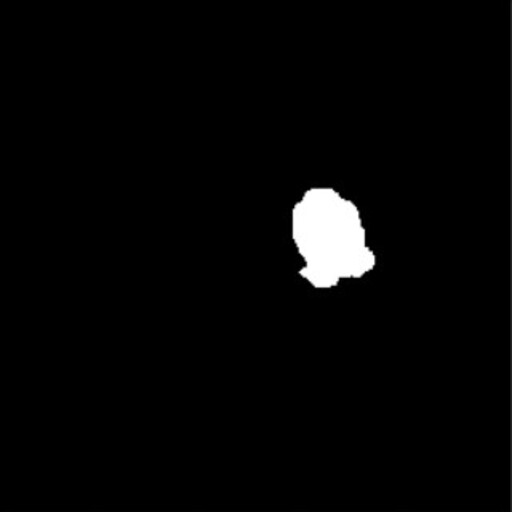} &
     \includegraphics[width=0.156\textwidth]{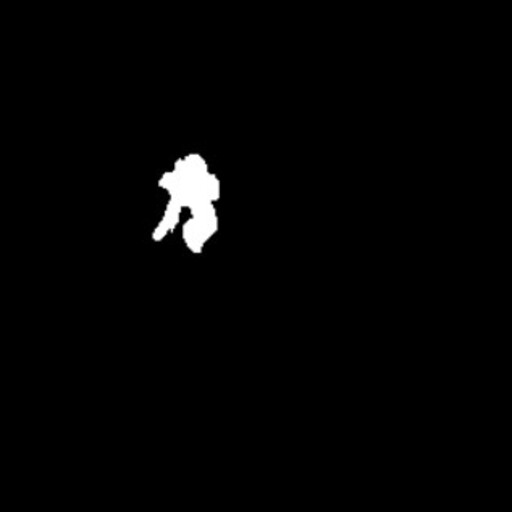}\\
     (d) & (e) & (f)

\end{tabular}}
\caption{Representative FLAIR 2D slices from the TCGA LGG dataset with the corresponding segmented ground truths. (d), (e) and (f) represent the corresponding ground truths for the FLAIR slices (a), (b), and (c) respectively.}
\vspace{-5pt}
\label{fig:TCGA_sample_images}
\end{figure}

The TCGA-LGG dataset acquired from The Cancer Imaging Archive (TCIA) {\cite{r18}} was used for training and validating the model on LGG slices exclusively. The dataset comprises a total of  3,929 images with corresponding binary segmentation masks. Of these, 1373 images had tumor regions and 2556 showed normal brain tissue. Each image has three channels- T1, FLAIR, T2, and T1-ce. However, several slices were missing the T1 and T1-ce channels. In such cases, the FLAIR modality replaced the missing channels. Some of the representative MR slices from the TCGA LGG dataset, along with their corresponding ground truths, is shown in Fig. \ref{fig:TCGA_sample_images}.
\vspace{-5pt}
\subsection{Methodology}
\label{sec:workflow}
\begin{figure*}[!h]
    \centering
     \includegraphics[width=\textwidth]{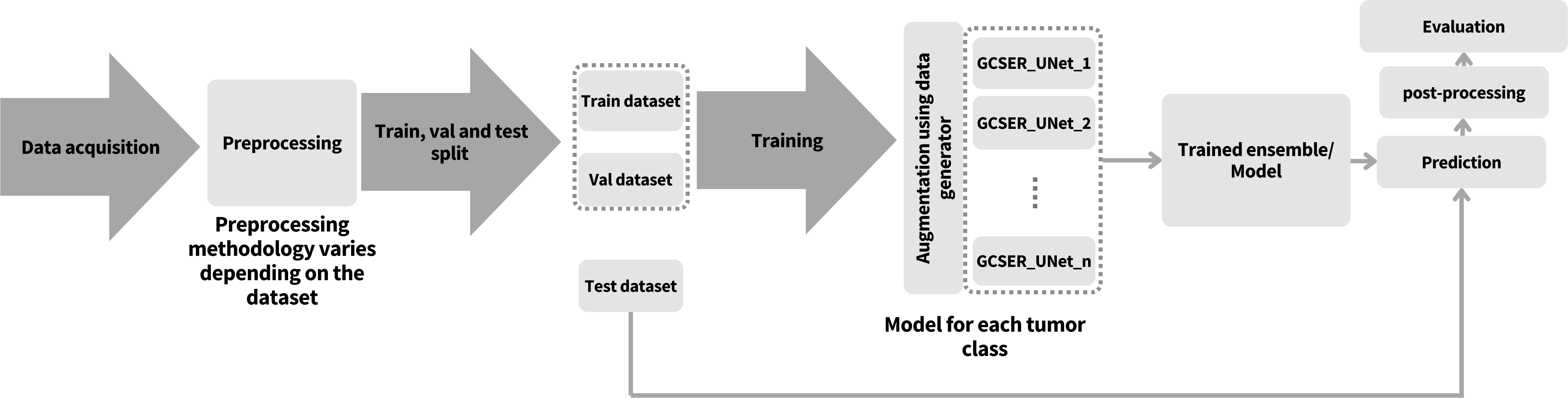}
    \caption{ Schematic representation of the methodology workflow.}
    \vspace{-7pt}
    \label{fig:workflow}
\end{figure*}

An overview of the methodology workflow for the implementation of the proposed method is provided in this section. Fig. \ref{fig:workflow} is a schematic representation of the workflow plan.
The following are the steps:
\begin{enumerate}
\item Pre-processing. Due to the two datasets' varied natures, two distinct pre-processing approaches have been adopted.
\item Splitting processed data into train test and validation datasets.
\item Training models for each tumor class while introducing augmentation via a data generator. 
\item Generation of test set segmentation outputs using trained model/ensemble.
\item Thresholding generated outputs and evaluation.
\end{enumerate}

The following subsections go into great depth about the implementation of these steps.
\subsubsection{Preprocessing}
\begin{figure*}[!h]
\setlength\tabcolsep{2pt}
\centering
\begin{tabular}{c c }
     
     \includegraphics[width=0.80\textwidth]{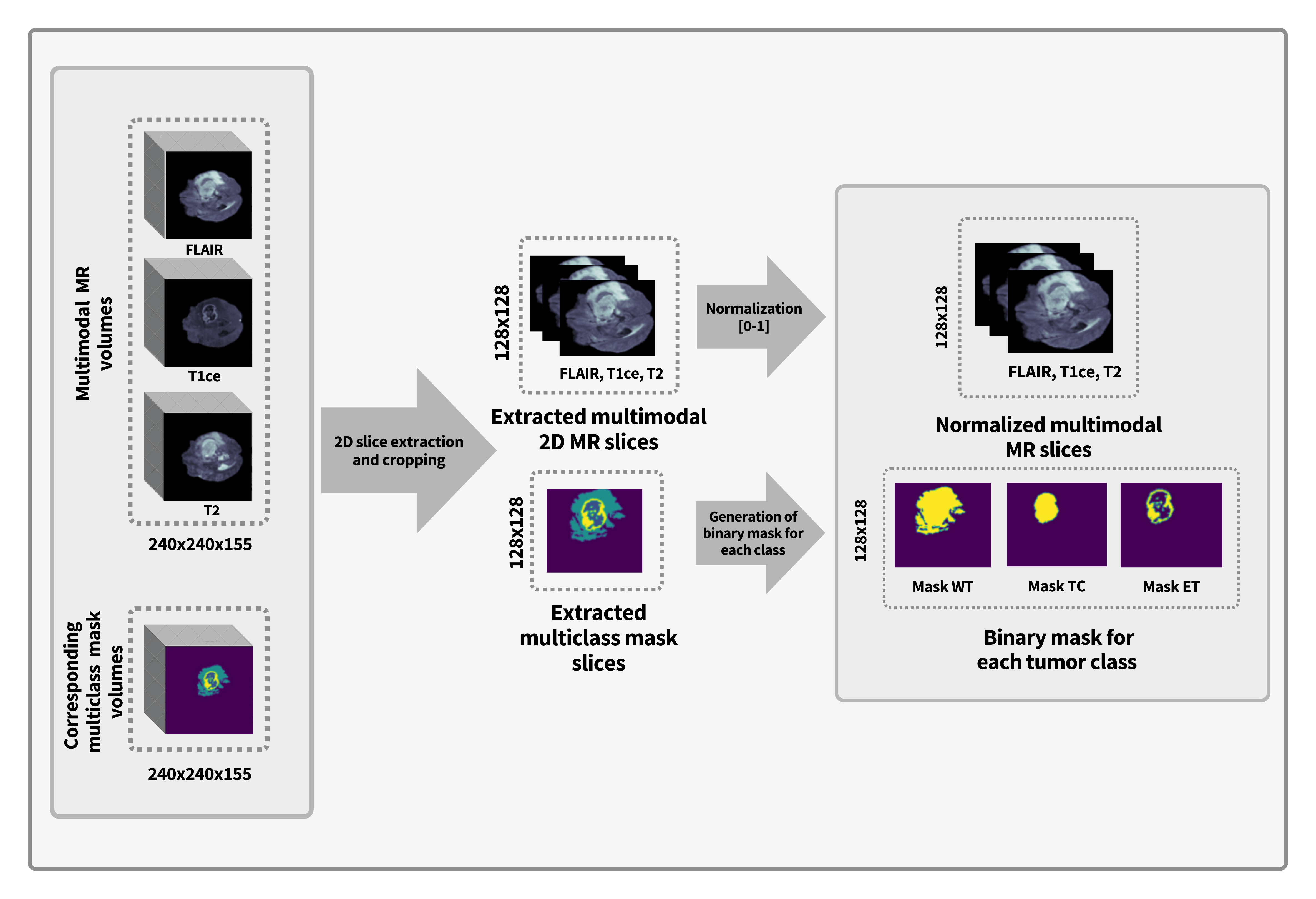} &
     \includegraphics[width=0.20\textwidth]{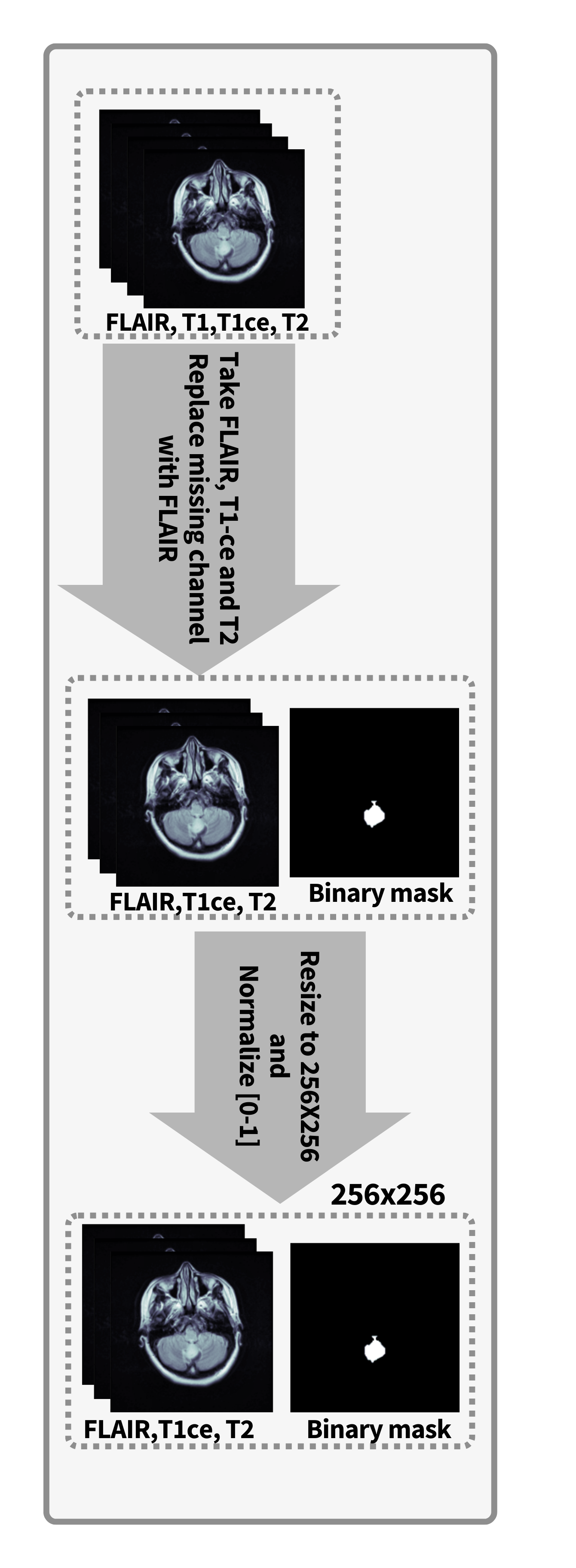}\\
      (a) & (b) \\

\end{tabular}
\caption{Pre-processing methodology. (a) Preprocessing methodology for the BraTS 2020 dataset. (b) Preprocessing methodology for the TCGA LGG dataset}
\label{fig:preprocessing}
\end{figure*}

From the BraTS 2020 training dataset, 2D slices were derived from each multimodal MR and their corresponding mask volumes. Slices containing at least one positive pixel in the corresponding mask slice were retained for subsequent processing, while the rest were discarded. The extracted 2D MR and mask slices were then cropped to dimensions 128$\times$128, following which the multiclass masks were modified to represent the tumor classes W, T, and E, respectively, before training. The MRI slices of modalities- FLAIR, T1-ce, and T2 were then merged to form 3-channel images. Each image and corresponding mask pair for the respective classes were then normalized to [0-1] and fed to the corresponding models for training. Figure \ref{fig:preprocessing}(a) shows a schematic illustration of this technique.

The TCGA LGG dataset consists of 2D MR slices of modalities T1, T1-ce, T2, and FLAIR, along with their binary masks representing class whole tumor (W). The FLAIR, T1-ce, and T2 MR slices are combined into a single 2D multichannel image. In case of missing MR modalities, they were replaced with the FLAIR modality. Finally, the multichannel MR images and their corresponding binary masks were normalized to [0-1] and resized to dimensions 256$\times$256$\times$3 before training. Fig. \ref{fig:preprocessing}(b) shows a schematic illustration of this technique.

\subsubsection{The proposed main architecture}
\begin{figure*}[h]
    \centering
    \begin{subfigure}{0.4\textwidth}
        \centering
        \includegraphics[width=\linewidth]{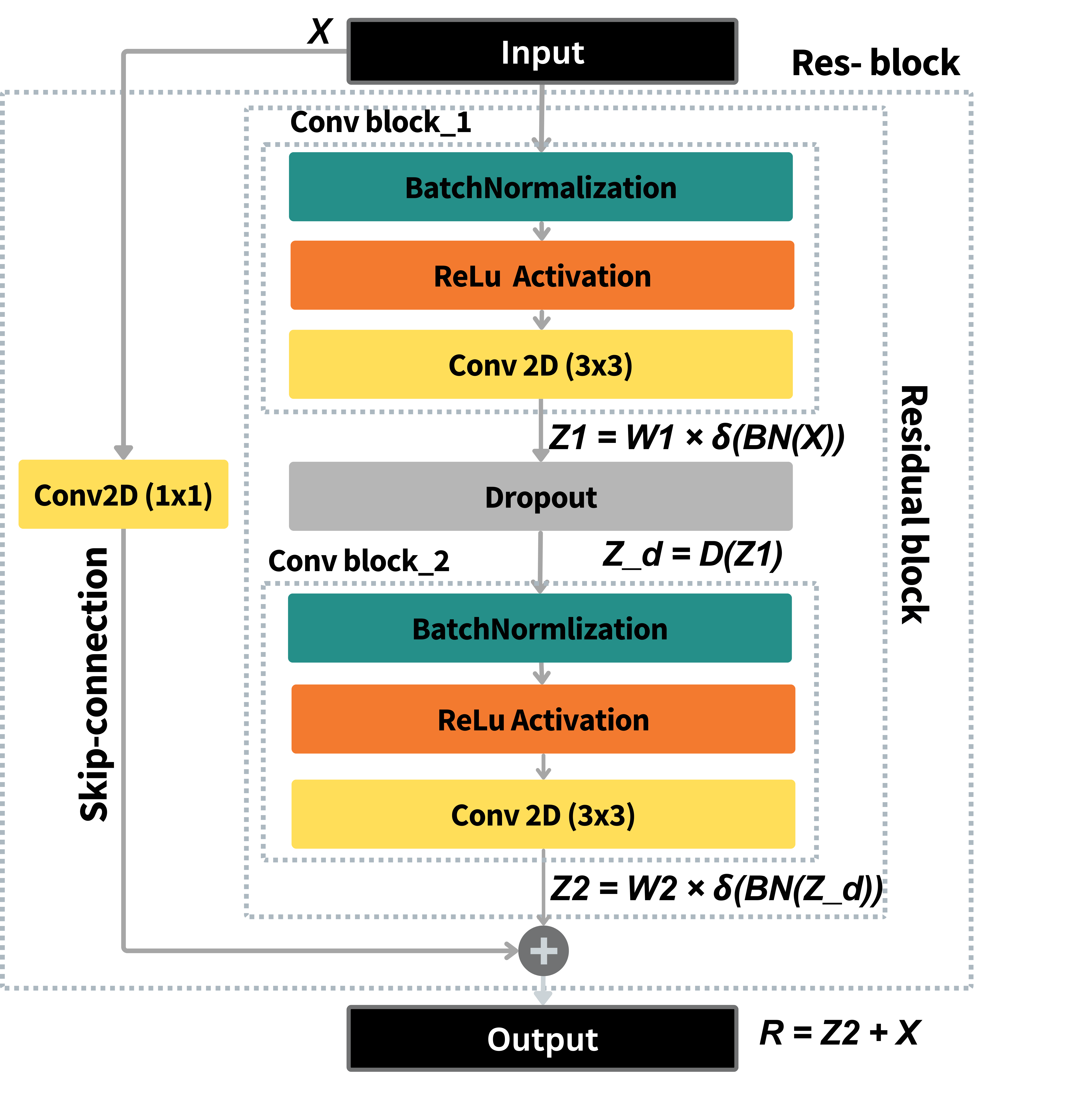}
        \caption{}
        \label{fig:Res_block}
    \end{subfigure}
    \hfill
    \begin{subfigure}{0.55\textwidth}
        \centering
        \includegraphics[width=\linewidth]{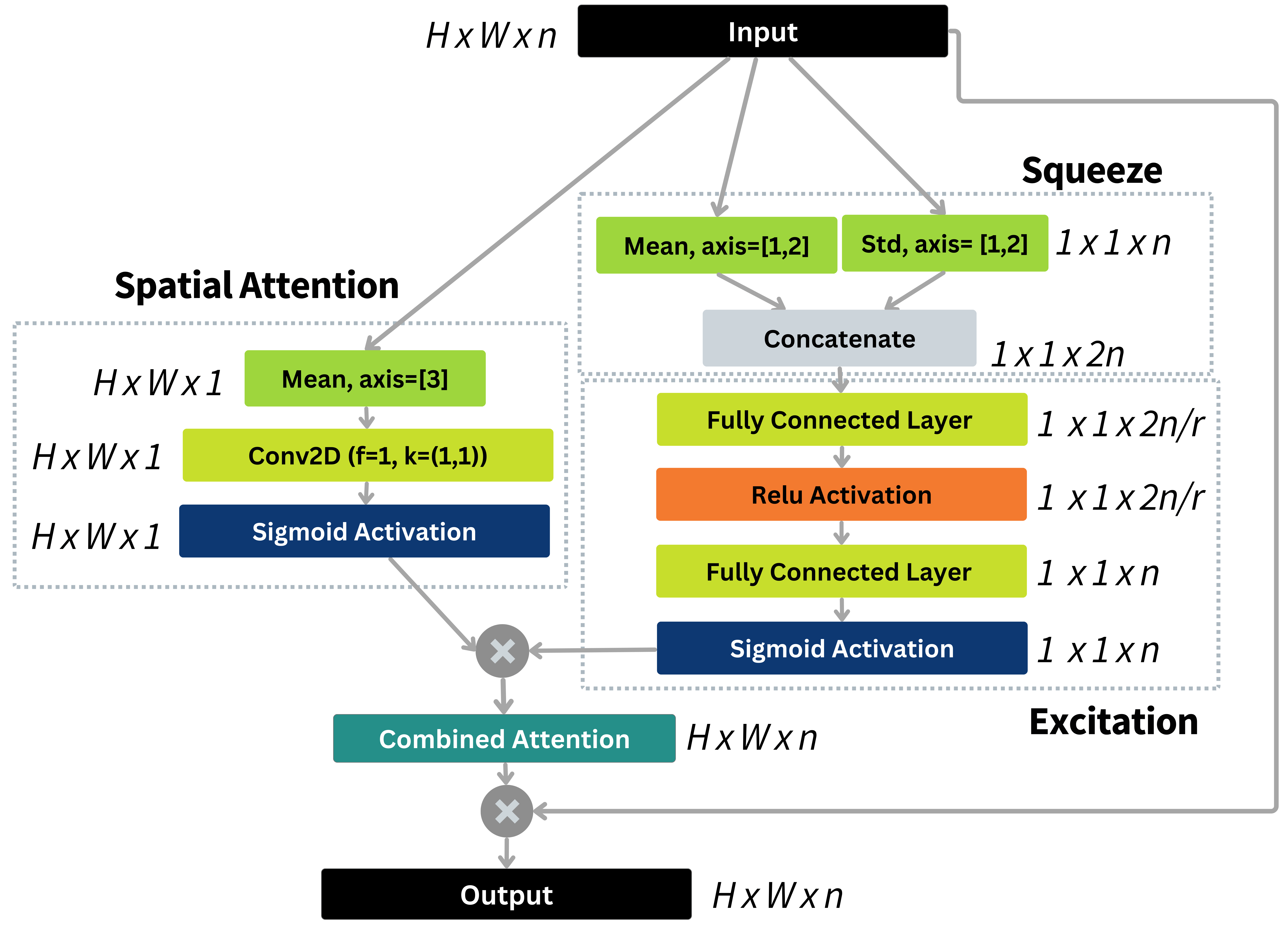}
        \caption{}
        \label{fig:GCSE_block}
    \end{subfigure}
    \par\medskip
    \begin{subfigure}{\textwidth}
        \centering
        \includegraphics[width=\linewidth]{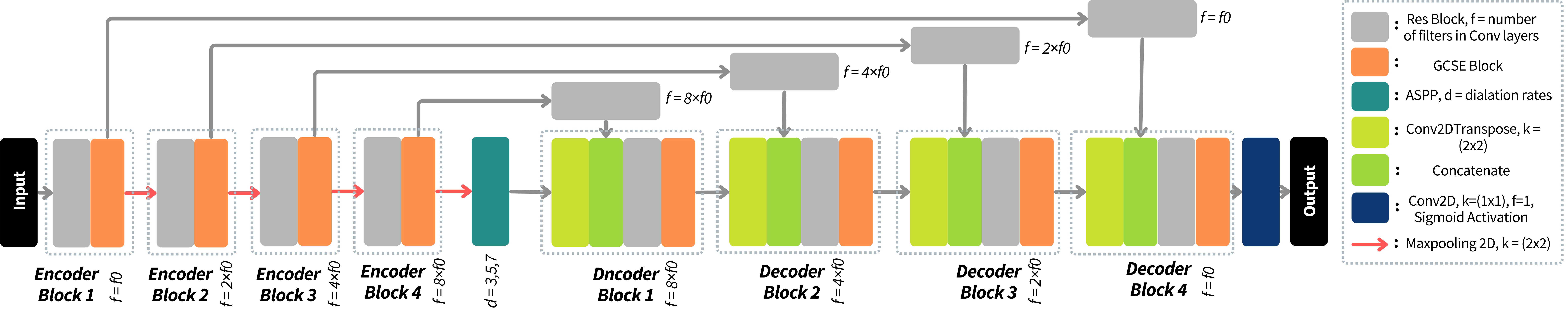}
        \caption{}
        \label{fig:model_architecture}
    \end{subfigure}
    \caption{(a) Schematic of Res-Block, (b) Schematic of the proposed GCSE mechanism, (c) Structural overview of the proposed GCSER-UNet}
    \label{fig:three_images}
\end{figure*}

The proposed network derives its name "GCSER-UNet" based on the repeated use of \textbf{s}queeze and \textbf{e}xcite blocks and employment of the \textbf{r}esidual learning approach at both the encoder and decoder. Fig. 5 shows the overall architecture of the proposed model. Fig. \ref{fig:Res_block} shows the Res block architecture, whereas Fig. \ref{fig:GCSE_block} shows the schematic of the proposed GCSE mechanism. Furthermore, Fig. \ref{fig:model_architecture} shows the schematic illustration of the encoder and decoder model architecture. Each of these architecture are thoroughly discussed in the subsections that follow.\\

\paragraph{Res block}
\label{sec:Res_block}
The residual block and its skip connection have been termed the Res-block. The Res-block has been implemented as an integral architectural aspect of the proposed architecture's contractive and expansive paths. Fig. \ref{fig:Res_block} presents a schematic representation of the sequential layers within a Res-block. The residual network in each Res-block incorporates two sequential Conv2D layers, each preceded by Batch-Normalization and ReLU activation layers. The Dropout layer randomly weights node outputs from hidden layers to $0$ with a predetermined rate ‘$r$’ (here $0.1$) at each training step, which builds resilience to overfitting. Node outputs not set to $0$ are scaled up by 1/(1 - $r$) such that the sum over all inputs remains constant. From Fig \ref{fig:Res_block} we can derive Equations \ref{eqn:e1} and \ref{eqn:e2}, which represent the outputs of the initial and subsequent convolution blocks within the 'Res' block. The operations BN(), $\delta$(), and D() correspond to Batch Normalization, Relu Activation, and Dropout, respectively.
\vspace{-3pt}
\begin{align}
    Z_1(X) &= W_1\times\delta(BN(X))\label{eqn:e1}\\
    Z_2(X) &= W_2\times\delta(BN(Z_d)),\hspace{0.07cm} where \hspace{0.07cm}Z_d = D(Z_1) \label{eqn:e2}\\
    R(X) &= Z_2(X) + X\label{eqn:e3}\\
    Z_2(x) &= R(X) - X\label{eqn:e4}
\end{align}
 Equation \ref{eqn:e3} signifies the final output of the 'Res' block. Rearranging Equation \ref{eqn:e3} demonstrates that $Z_2(X)$ can be depicted as the difference between the desired output $R(X)$ and the input $(X)$, termed as the residue. Kiaming et al. argue that learning this residue is comparatively easier. This can be justified by considering that, for $R(x)$ to represent the identity function $X$, the residue $Z_2(X)$ must be nullified, necessitating the weight matrix $W2$ to approach zero. Conversely, in the absence of the skip connection $(R(x) = Z_2(x))$, adjustments to the weights and bias values become imperative to adhere to the identity function. Learning an identity function from scratch for a non-residual network is notably challenging, exacerbated by the non-linearity within the layers, leading to the degradation problem. Table \ref{tab:table_1} discusses the sequence of layers in the Res-block in more detail. Here, $f_i$ denotes the number of filters in the $i^{th}$ encoder block, $k$ represents the kernel size, $d$ signifies the dropout percentage, and $s$ corresponds to the stride. Zero padding was applied for each convolution to maintain consistent input and output feature map dimensions.

\begin{table}[!h]
    \centering
    \caption{Sequence of layers in the Res-block, where layer 8 represents the skip connection.}
\resizebox{0.47\textwidth}{!}{\begin{tabular}{|c|c|c|}
        \hline
       Layer no. & Layer type & connected to layer \\
        \hline
         1 & BatchNormalization & input\\
         2 & ReLU Activation & 1\\
         3 & Conv2D($f_i=2^{i-1}\times32,k=3,s=1$) & 2\\
         4 & Dropout(d=0.1) & 3\\
         5 & BatchNormalization & 4\\
         6 & ReLU Activation & 5\\
         7 & Conv2D($f_i=2^{i-1}\times32,k=3,s=1$) & 6\\
         8 & Conv2D($f_i=2^{i-1}\times32,k=1,s=1$) & input,9\\
         9 & Addition & 7, 8\\
        \hline
    \end{tabular}}
    \label{tab:table_1}
\end{table}
\paragraph{GCSE Block}
The GCSE block represents an advanced extension of the conventional Squeeze-and-Excitation (SE) block, designed to capture comprehensive global context information within neural networks, as shown in Fig. \ref{fig:GCSE_block}. Initially, in the Squeeze stage, the mean $(M_{(k)})$ and standard deviation $(S_{(k)})$ for the $k^{th}$ channel are computed from the input tensor $(X)$, as shown in equations \ref{eqn:SE_1} and \ref{eqn:SE_2}.

\vspace{-3pt}
\begin{align}
    M_{(k)} &= \frac{1}{H \times W}\sum_{i=1}^{H}\sum_{j=1}^{W} X(i,j,k)\label{eqn:SE_1}\\
    S_{(k)} &= \sqrt{\frac{1}{H \times W}\sum_{i=1}^{H}\sum_{j=1}^{W} (X(i,j,k) - M_{(k)}))^2}\label{eqn:SE_2}
\end{align}

These operations extract vital channel-wise statistics, enabling the model to understand essential channel dependencies. Subsequently, in the Excitation stage, the channel-wise attention $(A_c)$ is obtained using the equation \ref{eqn:SE_3}.

\vspace{-3pt}
\begin{align}
   A_c = \sigma(W_2 \delta(W_1 \begin{bmatrix} M \\ S \end{bmatrix}))\label{eqn:SE_3}
\end{align}

Where $\sigma$ and $\delta$ represent the sigmoid and ReLU activation functions, respectively. The parameter ratio $(r)$ controls the dimensionality reduction in the intermediate layers, impacting the expressiveness and computational complexity of the GCSE block. The incorporation of standard deviation in the GCSE block stems from observing elevated standard deviation in channels with notable contrast between abnormalities (e.g., tumors) and normal tissues. Simultaneously, the spatial attention $(A_s)$ is derived through equation \ref{eqn:SE_4}.

\vspace{-3pt}
\begin{align}
   A_{s} = \sigma(W_3 \delta(\frac{1}{n} \sum_{k=1}^{n} X(i, j, k)))\label{eqn:SE_4}
\end{align}
The combination stage integrates the channel-wise and spatial attention outputs, generating a comprehensive attention map $(A)$ given by equation \ref{eqn:SE_5}. Finally, the feature rescaling stage dynamically reweights the input features $(X)$ according to the attention map $(A)$, as given by equation \ref{eqn:SE_6}.
\begin{align}
   A = A_{c} \odot A_{s}\label{eqn:SE_5}\\
   Y = X \odot A\label{eqn:SE_6}
\end{align}
Through these intricately orchestrated mathematical operations, the GCSE block significantly enhances the network's ability to discern complex patterns and salient features within the data, thereby fostering an enriched understanding of global context and intricate relationships.

\paragraph{Encoder}
The encoder of the GCSER-UNet is predominantly a cascade of structurally repeating units known as encoder blocks, with each successive unit in the cascade having double the number of feature channels compared to its preceding counterpart. Each encoder block is composed of a Res-block followed by a GCSE block. Integrating the GCSE blocks significantly enhances model performance through dynamic feature recalibration. This recalibration facilitates nuanced channel-wise and spatial attention fusion, vital for capturing complex inter-channel dependencies and extracting essential global context information, thereby ensuring precise and accurate brain tumor segmentation.

The output from each encoder block follows two paths: the first path, known as the (long) skip connection, directs the output feature map through a Res-block and then to the corresponding decoder block. The second path subjects the output feature map to a MaxPooling2D operation with a stride of 2, effectively reducing its spatial dimensions by half before feeding it as input to the subsequent encoder block. The sequential downsampling and increase in the number of feature channels enhance the model's receptive field, facilitating the extraction of more contextual information.

The pooled output from the final encoder block undergoes processing via the ASPP module, incorporating atrous convolutions at multiple dilation rates and spatial pyramidal pooling. This integration allows the model to capture contextual information at different scales, a crucial aspect of brain tumor segmentation tasks. The proposed model utilizes four parallel 3$\times$3 convolutions with 3, 5, and 7 dilation rates, respectively. The image-level feature is generated using global average pooling. The features from all branches are upsampled to the input size using bilinear interpolation, which is then concatenated and passed through another 1$\times$1 Conv2D layer. The resulting output is then provided to the decoder.

\hspace{20 mm} 
\paragraph{Decoder}
Similarly, as the encoder, the decoder comprises structurally repeating units. There is a corresponding decoder block for each encoder block. Each decoder block produces a twofold increase in the size of the input feature map by performing a 2$\times$2 Conv2DTranspose operation on it. The upsampled output is then concatenated to the skip feature maps acquired from the corresponding encoder block. The resulting feature maps are sent through a Res-block (see section \ref{sec:Res_block}) followed by the GCSE unit. Ultimately, a 1$\times$1 Conv2D operation followed by sigmoid activation creates the final output.

\section{Experimental results}
\subsection{Implementation details}
The code was executed in TensorFlow version 2.6.4, and the models were trained on a Kaggle cloud-based server using an NVIDIA Tesla P100 GPU, CUDA version 11.8, cuDNN version 8.9.0.
\subsection{Experimental settings}
A 75:15:10 train-validation-test split ratio was employed for each dataset (All results presented in this paper are based on this test set). During the training process, image-data-generator was used to introduce augmentations such as random flip (horizontal and vertical), random rotate, random zoom, etc.. The Adam optimizer with an initial learning rate of 0.001 and \emph{reduce on plateau}(for val loss) with \emph{decay factor} = 0.2 and $patience$ = 5 was used for training the models. The models took 50 epochs to converge for both datasets.
\subsection{Evaluation metrics}
The model's effectiveness was assessed using the following metrics:\\

(1) \textbf{Dice coefficient}:
\\The Dice coefficient (Dice) is calculated by taking twice the intersection of the two sets and dividing it by the sum of their sizes. In image segmentation, the sets represent the pixels or voxels within the predicted and ground truth regions. Mathematically it can be represented as shown in eqn \ref{eqn:Dice_1}.
\begin{equation}\label{eqn:Dice_1}
    Dice =  {\frac{2\times|T{\cap}\overline{T}|}{|T{\cup}\overline{T}|}}\hspace{1cm} where, \hspace{0.15cm}Dice {\in} [0,1]
\end{equation}
Here, Class $T$: tumor, Class $\overline{T}$: non-tumor. It can also be represented as eqn \ref{eqn:Dice coeff}.
\begin{equation}
\label{eqn:Dice coeff}
    Dice =  {\frac{2{\times}Pos}{2{\times}Pos + \overline{Pos} + \overline{Neg}}}\hspace{0.15cm} where, \hspace{0.15cm}Dice {\in} [0,1]
\end{equation}
Here $Pos$ = instances from true positive class, $\overline{Pos}$ = instances from false positive class, $Neg$ = instances from true negative class, $\overline{Neg}$ = instances from false negative class.\\

(2) \textbf{IoU}: 
\\IoU quantifies the extent of agreement or overlap between the predicted location of an object (determined by a machine learning model) and the ground truth location (manually labeled). It provides a measure of how well the predicted region aligns with the actual region of the object. It can be represented in equation \ref{eqn:IOU}.
\begin{equation}\label{eqn:IOU}
    IoU =  {\frac{|T{\cap}\overline{T}|}{|T{\cup}\overline{T}|}} = {\frac{Pos}{Pos + \overline{Pos} + \overline{Neg}}} \hspace{0.2cm} where, \hspace{0.05cm}IoU {\in} [0,1]
\end{equation}
where, Class $T$: tumor and Class $\overline{T}$: non-tumor.
\\(3) \textbf{Sensitivity:} 
\\The metric sensitivity measures how well a model can predict true positives for each provided class. It is represented in eqn \ref{eqn:sensitivity}.
\begin{equation}\label{eqn:sensitivity}
    Sensitivity = {\frac{Pos}{Pos + \overline{Neg}}} 
\end{equation}
\\(4) \textbf{Specificity:}
\\Specificity measures the ability of a model to predict true negatives for each class provided, as represented by the eqn \ref{eqn:specificity}.
\begin{equation}\label{eqn:specificity}
    Specificity = {\frac{Neg}{Neg + \overline{Pos}}} 
\end{equation}

(5) \textbf{Loss function}
The loss function $L_{FD}$ utilized in this work is a mix of Dice loss and Focal loss, as shown in eqn \ref{eqn:loss_function_1}.

\begin{equation}\label{eqn:loss_function_1}
    L_{FD} = L_{Dice} + L_{Focal}
\end{equation}
where $L_{Dice}$ and $L_{Focal}$ are Dice loss and Focal loss, respectively. \\
Dice loss may be defined as stated in eqn \ref{eqn:loss_function_4} and depends on the Dice coefficient.
\begin{equation}\label{eqn:loss_function_4}
    L_{Dice} = 1 - \text{Dice}
\end{equation}
Focal loss is a binary cross-entropy loss version that tackles the class imbalance issue with the conventional cross-entropy loss by down-weighting the contribution of easy-to-categorize samples, allowing learning of tougher cases. It is defined in eqn \ref{eqn:loss_function_2}.
\begin{equation}\label{eqn:loss_function_2}
    L_{Focal(P_{T})} = \alpha(1-E_{t})^\gamma \cdot L_{BCE(P,y)}
\end{equation}
Here, $L_{BCE}$ is the Binary Cross Entropy loss. The likelihood of correctly guessing the ground truth class, $E_{t}$, is defined in equation \ref{eqn:loss_function_3}.
\begin{equation}\label{eqn:loss_function_3}
   E_{t} = 
\begin{cases}
    E,    & \text{if } y=1 \\
    1-E,  & \text{if } y=0
\end{cases}
\end{equation}
The degree to which easy-to-classify cases are down-weighted so that learning the challenging instances can receive more attention is controlled by the parameters $\alpha$ and $\gamma$. For $\gamma$ = 0, the Focal loss is simplified to the Binary Cross Entropy loss.
\begin{figure}[!hbt]
        \centering
        \includegraphics[width=0.45\textwidth]{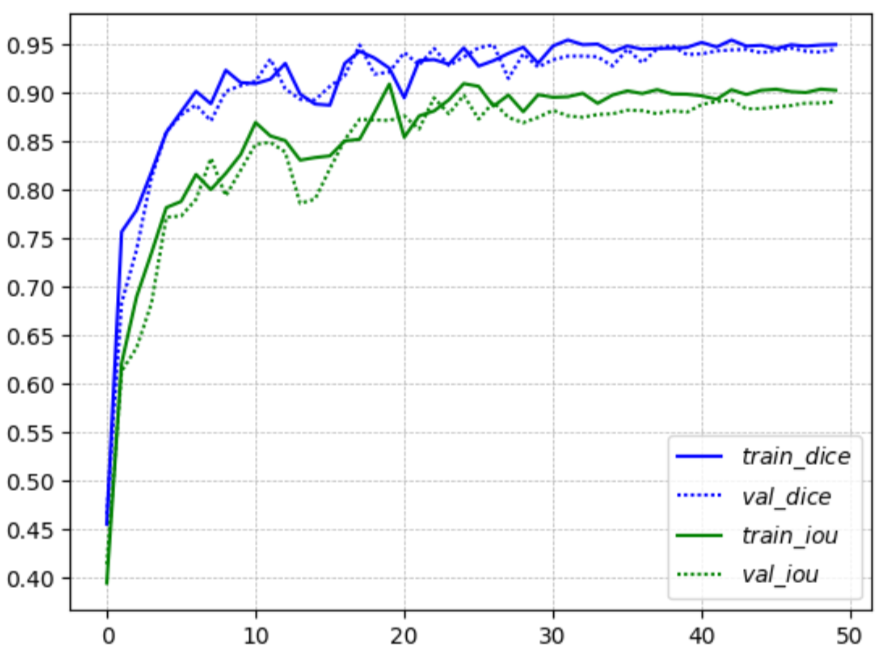}
        \caption{Training-validation curves of Dice Coefficient and IoU on TCGA LGG dataset }
        \vspace{-3pt}
        \label{fig:TCGA dice_iou curve}
\end{figure}
\subsection{Ablation Results}
This subsection presents the ablation study results performed on TCGA LGG and BraTS 2020 datasets. Its main objective is to assess the impact of the specific components used in the proposed architecture, leading to improved segmentation performance. The ablation study was conducted on the test sets using the following configurations: (i) U-Net (ii) Res-UNet, (iii) SE-Res-UNet, (iv) SE-Res-UNet + ASPP, and (v) the proposed GCSER-UNet. 
\subsubsection{Performance on the TCGA LGG dataset}
Among the 1373 subjects exhibiting positive mask values, a selection of 1098 subjects was utilized for training purposes. Additionally, 138 subjects were allocated for validation, while the remaining 137 were reserved for testing. 

\begin{figure*}[h!]
    \centering
    \includegraphics[width=0.95\textwidth]{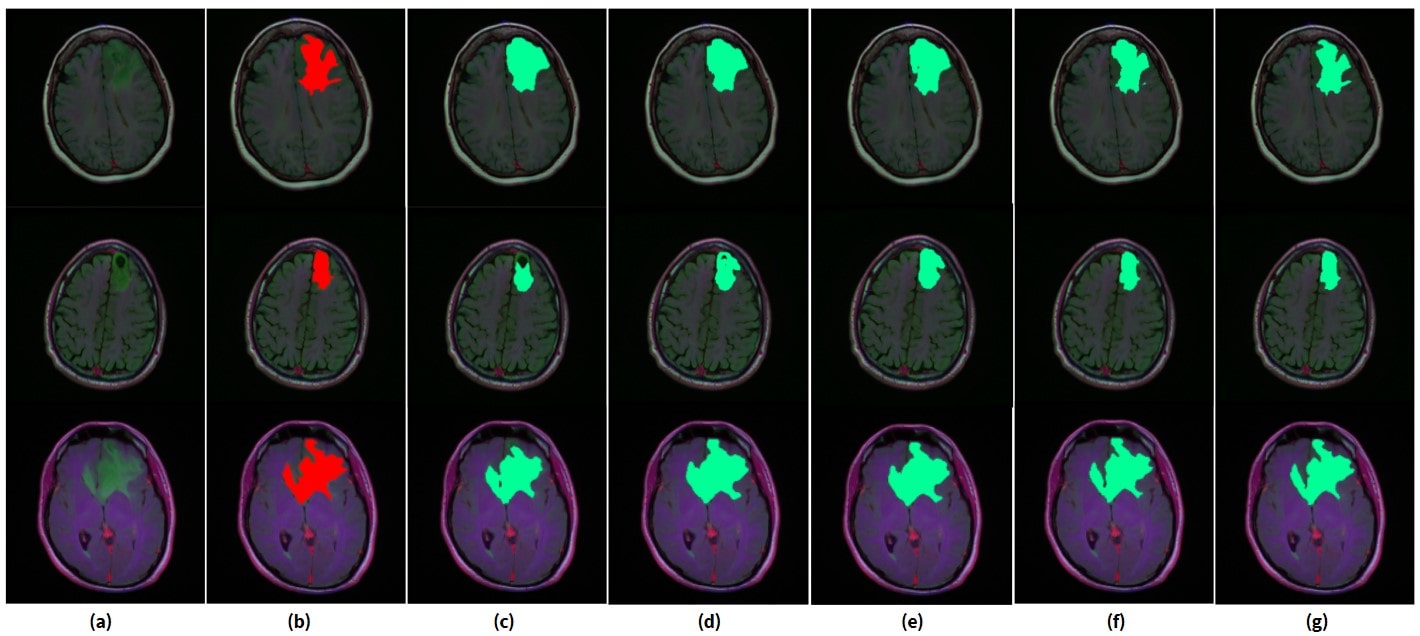}
    \caption{A comparison of the segmentation results produced by the GCSER-UNet and different variants of the UNet on random slices from the TCGA LGG testing data. (a) Original FLAIR image, (b) FLAIR image with ground-truth, (c) Predicted mask: UNet, (d) Predicted mask: Res-UNet, (e) Predicted mask: SE-Res-UNet, (f) Predicted mask: SE-Res-UNet+ASPP, (g) Predicted mask: GCSER-UNet}
    \vspace{-7pt}
    \label{fig:TCGA_ablation_results}
\end{figure*}

Fig \ref{fig:TCGA dice_iou curve} shows the training-validation dice and IoU curves, which depict well-fitted curves. Table \ref{tab:ablation_TCGA} depicts the performance metrics value evaluated on the TCGA LGG dataset, such as dice, IoU, sensitivity, and specificity. SE-Res-UNet + ASPP obtained better results than U-Net, Res-UNet and SE-Res-UNet. However, the proposed GCSER-UNet outperformed all other combinations by obtaining 94\% Dice score, 89\% IoU score, 97.7\% sensitivity and 99.97\% specificity. Also, Fig \ref{fig:TCGA_ablation_results} shows the superiority of the proposed GCSER-UNet when the predicted masks obtained from the above-mentioned configurations are qualitatively compared with each other. GCSER-UNet achieved the best segmentation results concerning the ground truth images, thus surpassing all other configurations.

\begin{table}[!h]
\caption{Comparison of the proposed model's performance with various other U-Net architectural variants on TCGA LGG dataset testing data. }
\begin{center}
\resizebox{0.47\textwidth}{!}{\begin{tabular}{  |c|c|c|c|c|}
\hline
Model & Dice  & IoU  & Sensitivity & Specificity  \\ 
\hline

UNet & 0.87 & 0.81  & 0.911 & 0.9991 \\ 
Res-UNet & 0.89 & 0.835 & 0.933 & 0.9995 \\
SE-Res-UNet & 0.91 & 0.85  & 0.957 & 0.9996\\ 
SE-Res-UNet + ASPP & 0.93 & 0.86 & 0.963 & 0.9997 \\ 
\textbf{Proposed GCSER-UNet} & \textbf{0.94}  & \textbf{0.88} & \textbf{0.977} & \textbf{0.9997}\\ \hline

\end{tabular}}
\end{center}
\label{tab:ablation_TCGA}
\end{table}

\begin{figure*}[hbt!]
\setlength\tabcolsep{2pt}
\centering
\resizebox{\textwidth}{!}{\begin{tabular}{c c c}
 \includegraphics[width=0.33\textwidth]{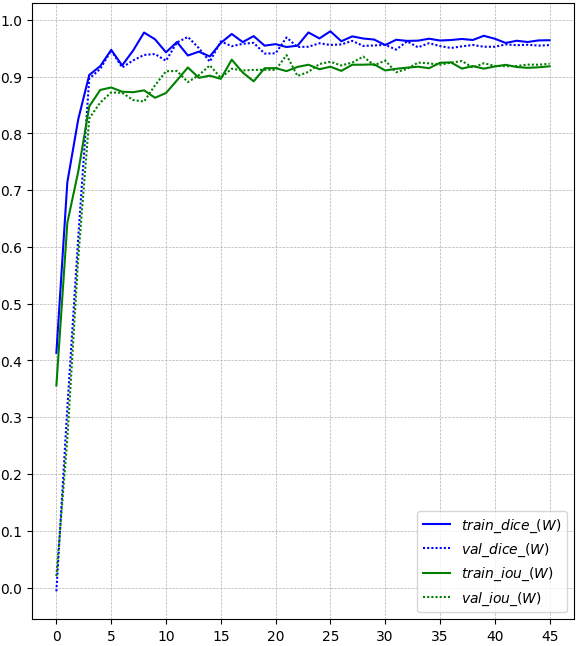} &
 \includegraphics[width=0.33\textwidth]{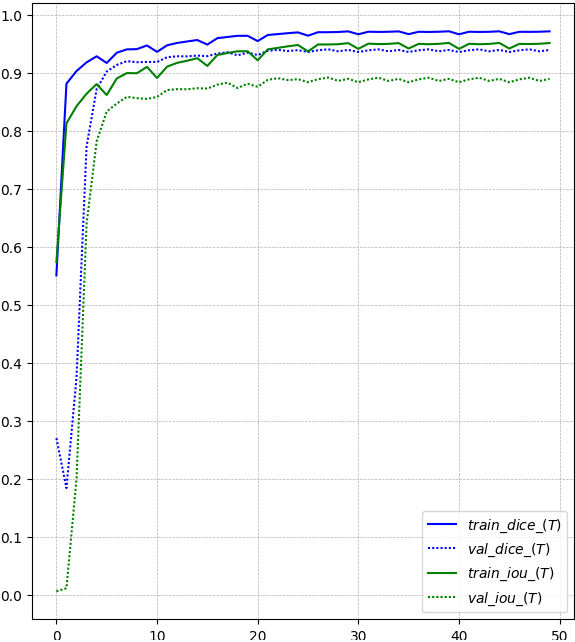} &
 \includegraphics[width=0.33\textwidth]{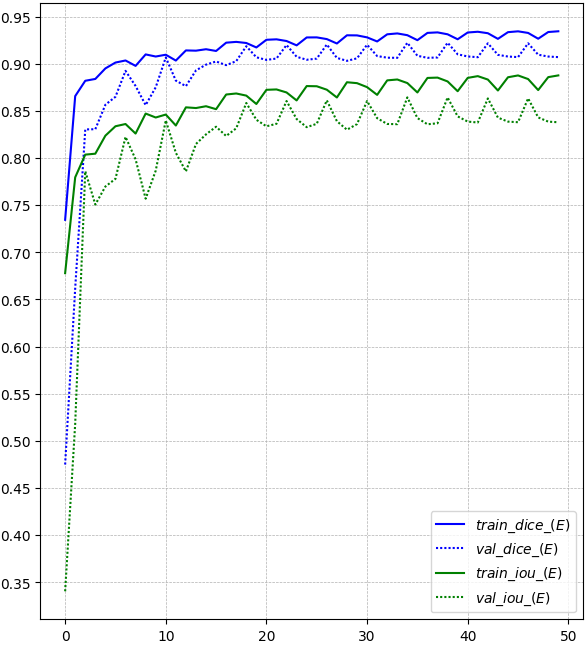} \\
 \textbf{(a)} &  \textbf{(b)} & \textbf{(c)} \\
\end{tabular}}
\caption{Traning validation curves of the metric Dice Coefficient and IoU for (a) GCSER-UNet$\_$W, (b) GCSER-UNet$\_$T, and (c) GCSER-UNet$\_$E for the BraTS 2020 dataset}
\vspace{-10pt}
\label{Brats_curves}
\end{figure*}

\vspace{-3pt}
 \begin{table*}[!hbt]
\caption{Performance evaluation of the proposed ensemble against several UNet architecture variations using test data from the BraTS 2020 dataset.}
\vspace{-5pt}
\begin{center}

\resizebox{0.95\textwidth}{!}{\begin{tabular}{| c | c  c  c | c  c  c | c  c  c | c  c  c| }
\hline
Architecture & \multicolumn{3}{ c |}{Dice} & \multicolumn{3}{ c |}{IoU} & \multicolumn{3}{| c |}{Sensitivity} & \multicolumn{3}{| c |}{Specificity}  \\ 
\cline{2-13}
& W & T & E & W & T & E & W & T & E & W & T & E \\
\hline

U-Net & 0.86 & 0.83 & 0.80 & 0.82 & 0.80 & 0.79 & 0.93 & 0.89 & 0.87 & 0.9991 & 0.9993& 0.9992\\ 
Res-UNet & 0.88 & 0.85 & 0.83 & 0.82 & 0.82 & 0.82 & 0.94 & 0.91 & 0.89 & 0.9992 & 0.9995& 0.9993 \\ 
SE-Res-UNet & 0.91 & 0.88 & 0.85 & 0.86 & 0.85 & 0.84 & 0.96 & 0.94 & 0.91 & 0.9994 & 0.9995 & 0.9991 \\ 
SE-Res-UNet + ASPP & 0.92 & 0.89 & 0.88 & 0.88 & 0.86 & 0.85 & 0.97 & 0.96 & 0.93 & 0.9993 & 0.9994 & 0.9992 \\
\textbf{Proposed GCSER-UNet} & \textbf{0.95} & \textbf{0.92} & \textbf{0.90} & \textbf{0.91} & \textbf{0.90} & \textbf{0.87} & \textbf{0.98} & \textbf{0.98} & \textbf{0.95} & \textbf{0.9997} & \textbf{0.9997} & \textbf{0.9997} \\ \hline

\end{tabular}}
\vspace{-10pt}
\label{tab:table_3}
\end{center}
\end{table*}
\vspace{-5pt}
\subsubsection{Performance on the BraTS 2020 dataset}
Slices from the 369 labeled subjects in the BraTS 2020 dataset were extracted as 2D images and masks. For each class, only slices with positive mask values were considered. 

\begin{figure*}[!h]
    \centering
    \includegraphics[width=0.95\textwidth]{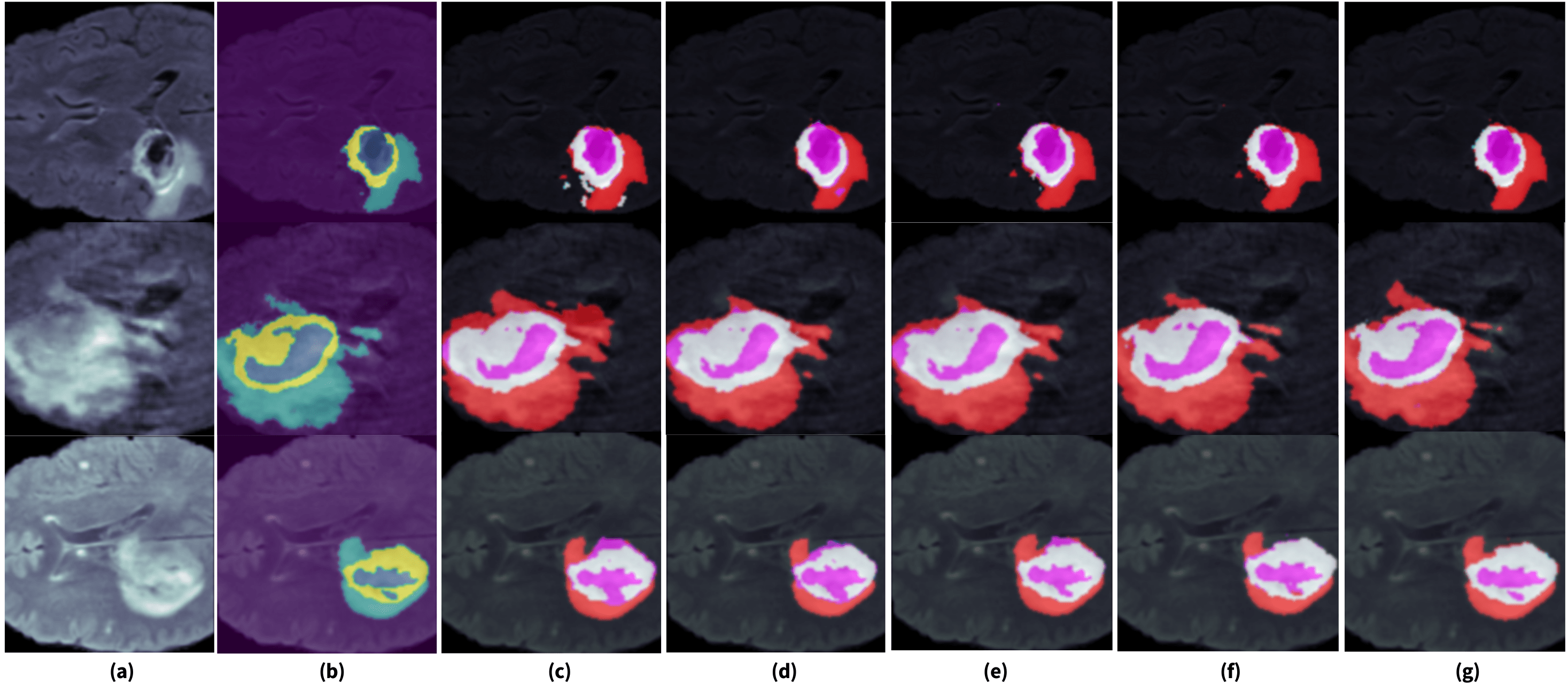}
    \caption{A comparison of the segmentation results of the GCSER-UNet and different variants of the UNet on random slices from the BraTS 2020 testing data. (a) Original FLAIR image, (b) FLAIR image with ground-truth, (c) Predicted mask: UNet, (d) Predicted mask: Res-UNet, (e) Predicted mask: SE-Res-UNet, (f) Predicted mask: SE-Res-UNet+ASPP, (g) Predicted mask: GCSER-UNet}
    \vspace{-15pt}
    \label{fig:ablation_results_BraTS}
\end{figure*}

Figure \ref{Brats_curves} illustrates the training-validation dice and IoU score curves on the BraTS 2020 dataset, demonstrating a strong convergence between the training and validation curves, thus indicating the model's high generalizability. Meanwhile, the findings from the ablation study based on the BraTS 2020 dataset testing data are detailed in Table \ref{tab:table_3}. Qualitative segmentation results from the ablation study are depicted in Figure \ref{fig:ablation_results_BraTS}. Table \ref{tab:table_3} distinctly highlights the improvement in segmentation output resulting from the architectural enhancements made to the original UNet framework.

\section{Discussions}
This section presents an evaluation of the effectiveness of the
GCSER-UNet model in comparison to state-of-the-art techniques. The significant outcomes and implications
resulting from the proposed model are outlined. It is essential to note that all comparisons presented in this section are based on the outcomes reported on their respective local test sets by the other authors.
The performance comparison of the suggested model has been laid out concerning both the TCGA LGG dataset and the BraTS dataset.
\subsubsection{TCGA LGG dataset}
A comprehensive comparative analysis assessed the suggested method's efficacy compared to several state-of-the-art techniques, as reported in Table \ref{tab:comparison_TCGA}. The results indicate that the GCSER-UNet model achieved the highest dice score of 94$\%$ and mean IOU score of 88$\%$, surpassing the performance of existing techniques.
\begin{table}[!h]
\caption{Results using the TCGA LGG test set for the suggested method and comparison with other state-of-the-art techniques.}
\vspace{-9pt}
\begin{center}
\resizebox{0.30\textwidth}{!}{\begin{tabular}{|l|c|c|}
\hline
Method & Dice  &  IoU   \\ 
\hline
Santosh et al.\cite{santosh2023resunet} & 0.905 & 0.829 \\
Buda et al.\cite{buda2019association} & 0.915 & 0.840 \\
Venkata et al.\cite{ventakasubbu2021deep} & 0.870 & 0.780 \\
Naser et al.\cite{naser} & 0.905 & 0.829 \\
Sourodip et al.\cite{r17} & 0.918 & 0.826 \\
\textbf{Proposed GCSER-UNet} & \textbf{0.94}  & \textbf{0.88} \\ \hline

\end{tabular}}
\end{center}
\label{tab:comparison_TCGA}
\end{table}

\subsubsection{BraTS 2020 dataset}

The trained ensemble was evaluated against some of the state-of-the-art techniques, and the comparative findings have been enlisted in Table \ref{tab:table_4}. The proposed GCSER-UNet ensemble method obtained the best mean dice score of 95$\%$, 92$\%$, and 90$\%$ for tumor subcategories-W, T, and E, respectively.

\begin{table}[!h]
\vspace{-5pt}
\caption{Comparison of the results of the proposed ensemble (GCSER-UNet$\_$W, GCSER-UNet$\_$T, and GCSER-UNet$\_$E) with some of the most advanced techniques on the local test set of the BraTS 2020 training dataset}
\vspace{-6pt}
\begin{center}

\resizebox{0.50\textwidth}{!}{\begin{tabular}{| l | c |  c  c  c |  }
\hline
Method & Dataset & \multicolumn{3}{ c |}{Dice}   \\ 
\cline{3-5}
&  & W & T & E\\
\hline

Sundaresan et al.\cite{r14} & BraTS 2020 & 0.89 & 0.85 & 0.83 \\ 
Yanwu Xu et al.\cite{r15} & BraTS 2017 & 0.87 & 0.78 & 0.77  \\ 
Varghese et al.\cite{r16} & BraTS 2017 & 0.84 & 0.84 & 0.77 \\
Ding et al.\cite{stack_ding_2019} & BraTS 2015 & 0.83 & 0.67 & 0.59 \\
Zhang et al.\cite{zhang_attention_2020} & BraTS 2017 & 0.87 & 0.77 & 0.72 \\
Ilyas et al.\cite{ilyas_2022_hybrid} & BraTS 2018 & 0.88 & 0.76 & 0.65\\
Ballestar et al.\cite{ballestar_2021} & BraTS 2020 & 0.85 & 0.85 & 0.77 \\
Findon et al.\cite{fidon2021generalized} & BraTS 2020 & 0.91 & 0.84 & 0.77\\
Aboelenein et al.\cite{aboelenein2022irdnu} & BraTS 2020 & 0.87 & 0.84 & 0.80 \\
(nn$\_$UNet) Hou et al.\cite{nn_UNet} & BraTS 2020 & 0.94 & \textbf{0.93} & 0.88 \\
\textbf{Proposed GCSER-UNet ensemble} & \textbf{BraTS 2020} & \textbf{0.95} & \textbf{0.92} & \textbf{0.90}\\ \hline

\end{tabular}}
\vspace{-15pt}
\label{tab:table_4}
\end{center}
\end{table}

\subsubsection{Major findings of the proposed architecture}
The following are the outlines of the most important findings obtained from the proposed model:
\begin{enumerate}
\item A single GCSER-UNet model was trained for the binary segmentation of brain tumors belonging to the TCGA LGG dataset. In contrast, an ensemble of three GCSER-UNets has been used for the multiclass segmentation of brain tumors on the BraTS 2020 dataset.

\item The suggested architecture achieved dice scores higher than many of the state-of-the-art 3D approaches without relying on any 3D context from the data, indicating that effective volumetric segmentation can be achieved purely based on the planar context.

\item The performance gains result from the improvements made to the original UNet\cite{ronneberger2015u} framework. Several studies that improved the performance of the UNet \cite{ronneberger2015u} by introducing architectural improvements were examined. Taking cues from these works\cite{r14,r15,r16,r17}, a multitude of changes were introduced to the U-Net\cite{ronneberger2015u} to develop a model that caters effectively to the task of brain tumor segmentation. 

\item The diverse features of tumor subcategories (W, T, and E) manifest distinctly in MR slices acquired from different modalities. The GCSE blocks introduce channel-wise weighting to optimize segmentation results, ensuring that channels with heightened tumor-related information substantially influence the output feature map. Simultaneously, the spatial attention mechanism facilitates effective discrimination between tumor and non-tumor tissue, enhancing the model's segmentation accuracy.

\item The Residual building blocks were used in both the contractive and the expansive path to tackle the degradation problem.

\item Taking into account the fact that brain MR images not only have plenty of local specifics but also have an almost limitless macro target expansion, ASPP was integrated to extract the multiscale features.

\item The total parameter count for the proposed architecture is approximately 8.1 million. Each model within the ensemble, trained on the BRaTS 2020 dataset, required around 0.02 seconds for inference per slice. In comparison, the 3D UNet baseline, with 22.6 million parameters, took approximately 0.047 seconds, while the 2D UNet baseline, with 7.7 million parameters, took about 0.019 seconds for inference per slice. Likewise, with the TCGA LGG dataset, the proposed architecture demonstrated an inference time of approximately 0.019 seconds, while the 2D UNet baseline exhibited an inference time of 0.0188 seconds per slice. These results indicate that our approach offers notable computational efficiency. Although comparing inference times with other state-of-the-art techniques would have been beneficial, most of their papers lack information on the inference time.

\end{enumerate}

\subsubsection{Limitation of the proposed architecture}
Despite obtaining exemplary segmentation results, the 2D segmentation approach stands as one of the major limitations of the proposed work. A 3D segmentation model can address this issue by utilizing inter-slice context and enhancing the model's functionality.
\vspace{-10pt}
\section{Conclusion}\
\vspace{-1pt}

This paper introduces the application of the unique 2D CNN architecture, GCSER-UNet, for precise brain tumor segmentation. The proposed method exhibits notable enhancements over existing techniques, as evidenced by the achieved dice scores (DC) of 95$\%$, 92$\%$, and 90$\%$ for tumor subregions W, T, and E, respectively, on the BraTS 2020 dataset. Notably, these results surpass the current state-of-the-art benchmark DC values of 93.93$\%$, 92.82$\%$, and 88.37$\%$ (nn$\_$UNet \cite{nn_UNet}). Furthermore, when tested on the TCGA LGG dataset, the model demonstrated a remarkable DC value of 94.3$\%$, outperforming the current state-of-the-art benchmark score of 91.8$\%$. Thus, the proposed model facilitates precise feature extraction and brain tumor segmentation. While the proposed model demonstrates its efficacy in accurately segmenting high-grade (HG) and low-grade (LG) volumes, future research endeavors could explore integrating contextual information from multiple planes, which remains unexplored in this study. Despite the potential rise in processing costs, this avenue of research holds promise for further refining the segmentation effectiveness of the proposed approach, thus offering a viable strategy to address the limitations observed in this study.
\vspace{-9pt}
\section*{Acknowledgement}
All authors declare that they have no known conflicts of interest regarding competing financial interests or personal relationships that could have an influence or are relevant to the work reported in this paper.

\section*{Code availibility}
 All code and implementation details are in \url{https://github.com/Sourjya261/Brain_tumor_segmentation}, which will be publicly available post-publication.
 \vspace{-7pt}

\bibliographystyle{unsrt}
\bibliography{references.bib}

@article{r9-10,
  title={Deep learning based brain tumor segmentation: a survey},
  author={Liu, Zhihua and Tong, Lei and Chen, Long and Jiang, Zheheng and Zhou, Feixiang and Zhang, Qianni and Zhang, Xiangrong and Jin, Yaochu and Zhou, Huiyu},
  journal={Complex \& Intelligent Systems},
  volume={9},
  number={1},
  pages={1001--1026},
  year={2023},
  publisher={Springer}
}

@inproceedings{ronneberger2015u,
  title={U-net: Convolutional networks for biomedical image segmentation},
  author={Ronneberger, Olaf and Fischer, Philipp and Brox, Thomas},
  booktitle={Medical Image Computing and Computer-Assisted Intervention--MICCAI 2015: 18th International Conference, Munich, Germany, October 5-9, 2015, Proceedings, Part III 18},
  pages={234--241},
  year={2015},
  organization={Springer}
}

@article{santosh2023resunet,
  title={Brain tumor segmentation of the FLAIR MRI images using novel ResUnet},
  author={Kumar, P Santosh and Sakthivel, VP and Raju, Manda and Satya, PD},
  journal={Biomedical Signal Processing and Control},
  volume={82},
  pages={104586},
  year={2023},
  publisher={Elsevier}
}

@article{ilyas_2022_hybrid,
  title={Hybrid-DANet: An Encoder-Decoder Based Hybrid Weights Alignment With Multi-Dilated Attention Network for Automatic Brain Tumor Segmentation},
  author={Ilyas, Naveed and Song, Yoonguu and Raja, Aamir and Lee, Boreom},
  journal={IEEE Access},
  volume={10},
  pages={122658--122669},
  year={2022},
  publisher={IEEE}
}

@article{ottom2022znet,
  title={Znet: deep learning approach for 2D MRI brain tumor segmentation},
  author={Ottom, Mohammad Ashraf and Rahman, Hanif Abdul and Dinov, Ivo D},
  journal={IEEE Journal of Translational Engineering in Health and Medicine},
  volume={10},
  pages={1--8},
  year={2022},
  publisher={IEEE}
}

@article{aboelenein2022irdnu,
  title={IRDNU-Net: Inception residual dense nested u-net for brain tumor segmentation},
  author={AboElenein, Nagwa M and Songhao, Piao and Afifi, Ahmed},
  journal={Multimedia Tools and Applications},
  volume={81},
  number={17},
  pages={24041--24057},
  year={2022},
  publisher={Springer}
}

@article{r17,
  title={Improved U-Net architecture with VGG-16 for brain tumor segmentation},
  author={Ghosh, Sourodip and Chaki, Aunkit and Santosh, KC},
  journal={Physical and Engineering Sciences in Medicine},
  volume={44},
  number={3},
  pages={703--712},
  year={2021},
  publisher={Springer}
}

@inproceedings{ballestar_2021,
  title={MRI brain tumor segmentation and uncertainty estimation using 3D-UNet architectures},
  author={Ballestar, Laura Mora and Vilaplana, Veronica},
  booktitle={Brainlesion: Glioma, Multiple Sclerosis, Stroke and Traumatic Brain Injuries: 6th International Workshop, BrainLes 2020, Held in Conjunction with MICCAI 2020, Lima, Peru, October 4, 2020, Revised Selected Papers, Part I 6},
  pages={376--390},
  year={2021},
  organization={Springer}
}

@inproceedings{fidon2021generalized,
  title={Generalized wasserstein dice score, distributionally robust deep learning, and ranger for brain tumor segmentation: BraTS 2020 challenge},
  author={Fidon, Lucas and Ourselin, S{\'e}bastien and Vercauteren, Tom},
  booktitle={Brainlesion: Glioma, Multiple Sclerosis, Stroke and Traumatic Brain Injuries: 6th International Workshop, BrainLes 2020, Held in Conjunction with MICCAI 2020, Lima, Peru, October 4, 2020, Revised Selected Papers, Part II 6},
  pages={200--214},
  year={2021},
  organization={Springer}
}

@inproceedings{r14,
  title={Brain tumour segmentation using a triplanar ensemble of U-Nets on MR images},
  author={Sundaresan, Vaanathi and Griffanti, Ludovica and Jenkinson, Mark},
  booktitle={International MICCAI Brainlesion Workshop},
  pages={340--353},
  year={2021},
  organization={Springer}
}

@article{zhang_attention_2020,
  title={Attention gate resU-Net for automatic MRI brain tumor segmentation},
  author={Zhang, Jianxin and Jiang, Zongkang and Dong, Jing and Hou, Yaqing and Liu, Bin},
  journal={IEEE Access},
  volume={8},
  pages={58533--58545},
  year={2020},
  publisher={IEEE}
}

@inproceedings{r15,
  title={Multi-scale masked 3-D U-net for brain tumor segmentation},
  author={Xu, Yanwu and Gong, Mingming and Fu, Huan and Tao, Dacheng and Zhang, Kun and Batmanghelich, Kayhan},
  booktitle={International MICCAI Brainlesion Workshop},
  pages={222--233},
  year={2019},
  organization={Springer}
}

@article{stack_ding_2019,
  title={A stacked multi-connection simple reducing net for brain tumor segmentation},
  author={Ding, Yi and Chen, Fujuan and Zhao, Yang and Wu, Zhixing and Zhang, Chao and Wu, Dongyuan},
  journal={IEEE Access},
  volume={7},
  pages={104011--104024},
  year={2019},
  publisher={IEEE}
}

@article{brats_3,
  title={Identifying the best machine learning algorithms for brain tumor segmentation, progression assessment, and overall survival prediction in the BRATS challenge},
  author={Bakas, Spyridon and Reyes, Mauricio and Jakab, Andras and Bauer, Stefan and Rempfler, Markus and Crimi, Alessandro and Shinohara, Russell Takeshi and Berger, Christoph and Ha, Sung Min and Rozycki, Martin and others},
  journal={arXiv preprint arXiv:1811.02629},
  year={2018}
}

@inproceedings{r18,
  title={Squeeze-and-excitation networks},
  author={Hu, Jie and Shen, Li and Sun, Gang},
  booktitle={Proceedings of the IEEE conference on computer vision and pattern recognition},
  pages={7132--7141},
  year={2018}
}

@article{brats_2,
  title={High resolution global gridded data for use in population studies},
  author={Lloyd, Christopher T and Sorichetta, Alessandro and Tatem, Andrew J},
  journal={Scientific data},
  volume={4},
  number={1},
  pages={1--17},
  year={2017},
  publisher={Nature Publishing Group}
}

@article{aspp,
  title={Deeplab: Semantic image segmentation with deep convolutional nets, atrous convolution, and fully connected crfs},
  author={Chen, Liang-Chieh and Papandreou, George and Kokkinos, Iasonas and Murphy, Kevin and Yuille, Alan L},
  journal={IEEE transactions on pattern analysis and machine intelligence},
  volume={40},
  number={4},
  pages={834--848},
  year={2017},
  publisher={IEEE}
}

@inproceedings{r16,
  title={Automatic segmentation and overall survival prediction in gliomas using fully convolutional neural network and texture analysis},
  author={Alex, Varghese and Safwan, Mohammed and Krishnamurthi, Ganapathy},
  booktitle={International MICCAI Brainlesion Workshop},
  pages={216--225},
  year={2017},
  organization={Springer}
}

@inproceedings{residual,
  title={Deep residual learning for image recognition},
  author={He, Kaiming and Zhang, Xiangyu and Ren, Shaoqing and Sun, Jian},
  booktitle={Proceedings of the IEEE conference on computer vision and pattern recognition},
  pages={770--778},
  year={2016}
}

@article{brats_1,
  title={The multimodal brain tumor image segmentation benchmark (BRATS)},
  author={Bjoern, H Menze and Andras, Jakab and Stefan, Bauer and Jayashree, Kalpathy-Cramer and Keyvan, Farahani and Justin, Kirby and others},
  journal={IEEE Trans. Med. Imaging},
  volume={34},
  number={10},
  pages={1993--2024},
  year={2015}
}

@article{r8,
  title={A survey of deep learning for MRI brain tumor segmentation methods: Trends, challenges, and future directions},
  author={Krishnapriya, Srigiri and Karuna, Yepuganti},
  journal={Health and Technology},
  pages={1--21},
  year={2023},
  publisher={Springer}
}

@article{r6,
  title={Recent Progress in Nanomedicines for Imaging and Therapy of Brain Tumors},
  author={Hasan, Ikram and Roy, Shubham and Guo, Bing and Du, Shiwei and Wei, Tao and Chang, Chunqi},
  journal={Biomaterials Science},
  year={2023},
  publisher={Royal Society of Chemistry}
}

@article{r4,
  title={Survival in a consecutive series of 467 glioblastoma patients: Association with prognostic factors and treatment at recurrence at two independent institutions},
  author={Blakstad, Hanne and Brekke, Jorunn and Rahman, Mohummad Aminur and Arnesen, Victoria Smith and Miletic, Hrvoje and Brandal, Petter and Lie, Stein Atle and Chekenya, Martha and Goplen, Dorota},
  journal={PloS one},
  volume={18},
  number={2},
  pages={e0281166},
  year={2023},
  publisher={Public Library of Science San Francisco, CA USA}
}

@article{r1-2,
  title={Immunotherapy for brain metastases and primary brain tumors},
  author={Di Giacomo, Anna M and Mair, Maximilian J and Ceccarelli, Michele and Anichini, Andrea and Ibrahim, Ramy and Weller, Michael and Lahn, Michael and Eggermont, Alexander MM and Fox, Bernard and Maio, Michele},
  journal={European Journal of Cancer},
  volume={179},
  pages={113--120},
  year={2023},
  publisher={Elsevier}
}

@article{r3,
  title={Immune surveillance of brain metastatic cancer cells is mediated by IFITM1},
  author={She, Xiaofei and Shen, Shijun and Chen, Guang and Gao, Yaqun and Ma, Junxian and Gao, Yaohui and Liu, Yingdi and Gao, Guoli and Zhao, Yan and Wang, Chunyan and others},
  journal={The EMBO Journal},
  pages={e111112},
  year={2023}
}

@article{r7,
  title={Radiation necrosis or tumor progression? A review of the radiographic modalities used in the diagnosis of cerebral radiation necrosis},
  author={Mayo, Zachary S and Halima, Ahmed and Broughman, James R and Smile, Timothy D and Tom, Martin C and Murphy, Erin S and Suh, John H and Lo, Simon S and Barnett, Gene H and Wu, Guiyun and others},
  journal={Journal of Neuro-Oncology},
  pages={1--9},
  year={2023},
  publisher={Springer}
}

@article{buda2019association,
  title={Association of genomic subtypes of lower-grade gliomas with shape features automatically extracted by a deep learning algorithm},
  author={Buda, Mateusz and Saha, Ashirbani and Mazurowski, Maciej A},
  journal={Computers in biology and medicine},
  volume={109},
  pages={218--225},
  year={2019},
  publisher={Elsevier}
}

@article{ventakasubbu2021deep,
  title={Deep learning-based brain tumour segmentation},
  author={Ventakasubbu, Pattabiraman and Ramasubramanian, Parvathi},
  journal={IETE Journal of Research},
  pages={1--9},
  year={2021},
  publisher={Taylor \& Francis}
}

@inproceedings{naser,
  title={Vox2Vox: 3D-GAN for brain tumour segmentation},
  author={Cirillo, Marco Domenico and Abramian, David and Eklund, Anders},
  booktitle={Brainlesion: Glioma, Multiple Sclerosis, Stroke and Traumatic Brain Injuries: 6th International Workshop, BrainLes 2020, Held in Conjunction with MICCAI 2020, Lima, Peru, October 4, 2020, Revised Selected Papers, Part I 6},
  pages={274--284},
  year={2021},
  organization={Springer}
}

@inproceedings{nn_UNet,
  title={Diffraction Block in Extended nn-UNet for Brain Tumor Segmentation},
  author={Hou, Qingfan and Wang, Zhuofei and Wang, Jiao and Jiang, Jian and Peng, Yanjun},
  booktitle={International MICCAI Brainlesion Workshop},
  pages={174--185},
  year={2022},
  organization={Springer}
}

\end{document}